\documentclass{article}

\usepackage{arxiv}

\usepackage{mathtools}
\usepackage[utf8]{inputenc} %
\usepackage[T1]{fontenc}    %
\usepackage{hyperref}       %
\usepackage{url}            %
\usepackage{booktabs}       %
\usepackage{amsfonts}       %
\usepackage{nicefrac}       %
\usepackage{microtype}      %
\usepackage{lipsum}
\usepackage{graphicx}

\title{Morphological Skip-Gram: Using morphological knowledge to improve word representation}

\author{
  Flávio A. Oliveira Santos \\%\thanks{web link, etc} \\
  Center of Informatics \\
  Federal University of Pernambuco \\
  Recife, Brazil \\
  \texttt{faos@cin.ufpe.br} \\
   \And
  Hendrik Teixeira Macedo \\
  Department of Computer Science \\
  Federal University of Sergipe \\
  São Cristóvão, Brazil \\
  \texttt{hendrik@dcomp.ufs.br} \\
  \And
  Thiago Dias Bispo \\
  Center of Informatics \\
  Federal University of Pernambuco \\
  Recife, Brazil \\
  \texttt{tdb@cin.ufpe.br} \\
   \And
  Cleber Zanchettin \\
  Center of Informatics \\
  Federal University of Pernambuco \\
  Recife, Brazil \\
  \texttt{cz@cin.ufpe.br} \\
}

\begin{document}
\maketitle

\begin{abstract}
Natural language processing models have attracted much interest in the deep learning community. This branch of study is composed of some applications such as machine translation, sentiment analysis, named entity recognition, question and answer, and others. Word embeddings are continuous word representations, they are an essential module for those applications and are generally used as input word representation to the deep learning models. Word2Vec and GloVe are two popular methods to learn word embeddings. They achieve good word representations, however, they learn representations with limited information because they ignore the morphological information of the words and consider only one representation vector for each word. This approach implies that Word2Vec and GloVe are unaware of the word inner structure. To mitigate this problem, the FastText model represents each word as a bag of characters n-grams. Hence, each n-gram has a continuous vector representation, and the final word representation is the sum of its characters n-grams vectors. Nevertheless, the use of all n-grams character of a word is a poor approach since some n-grams have no semantic relation with their words and increase the amount of potentially useless information. This approach also increases the training phase time. In this work, we propose a new method for training word embeddings, and its goal is to replace the FastText bag of character n-grams for a bag of word morphemes through the morphological analysis of the word. Thus, words with similar context and morphemes are represented by vectors close to each other. To evaluate our new approach, we performed intrinsic evaluations considering 15 different tasks, and the results show a competitive performance compared to FastText. 
\end{abstract}


\section{Introduction}
Natural language processing (NLP) is a branch of study that helps computers understand, interpret, and manipulate human language allowing applications to read, hear, interpret it, measure sentiment and deal with unstructured data. Most of the knowledge generated today is unstructured, from medical records to social media and extract and understand these pieces of information demand intelligent models able to process all of this information.

Deep Learning (DL) models have achieved the state-of-the-art results in many NLP tasks as machine translation \cite{googlenmt}, question and answering (Q$\&$A) \cite{lu2016knowing}, image captioning \cite{lu2016knowing}, keywords extraction \cite{zhang2020keywords}, text summarization \cite{paulus2017deep}, named entity recognition \cite{juniorarquitetura} and sentiment analysis \cite{lakkaraju2014aspect}. Most of these solutions use Word Embeddings (WE) to represent the input as continuous word vectors. Each word has one-word embedding. WE are fundamental to achieve good results in NLP tasks.

Word embedding's methods primary goal is to learn good words representations to be given as input for the machine learning model. Thus, they need to represent the maximum word information as possible. Every word has semantic and syntactic information, such as synonyms, antonyms, radical, lemma, stemming, morphological knowledge, part-of-speech, and others. Although the word embeddings need to represent most word information as possible, its training can not be computationally costly because it will be used in an expert system. Thus, there is a trade-off between being informative and not be computationally expensive.

There are three main approaches to learn WE: (i) Context-window based models, (ii) Semantic relationship-based models, and (iii) Graph Distance-based models. Each method has drawbacks, though; models from (ii) and (iii) use knowledge bases such as WordNet \cite{wordnet} and Freebase \cite{freebase} to learn the WE, but consider only a tiny part of the dataset. Models from (iii) use mainly the Leacock-Chodorow \cite{lch} distance to capture the semantic relationship between two words, ignoring alternative graph distances. Finally, knowledge bases involved are often limited to specific fields. Some models from (i), such as Neural Language Model \cite{nlm} and Word2Vec \cite{word2vec}, despite the good results, are trained using only local context information of every word instead of global context information. The Word2Vec method also does not use the internal word structure information (Morphology information). Although GloVe \cite{glove} uses global word context information, words with different lexical but equal meaning (paraphrases) have different representations because they are in a diverse global context. The FastText \cite{bojanowski2016enriching} model, based on Word2Vec, proposes the use of the internal word structure based on a bag of all n-gram characters of each word. Besides being a brute-force solution, some character n-grams have no semantic relationship with the formed word; the word 'American' and their character n-gram 'erica', for instance, have no semantic connection at all. %

We propose the Morphological Skip-Gram (MSG) model, whose main goal is to replace the bag of characters n-grams used in FastText with a bag of morphemes obtained from word morphological analysis. A morpheme is the smallest grammatical unit in a language. Thereby, words with common morphemes will have a similar representation. Our approach is important because it allows the uses of the word inner structure that has a syntactic relation with the complete word. Considering grammatically well-behaved texts, our approach sounds like a more consistent scientific hypothesis than uses all characters n-gram (FastText). 

The rest of this paper is organized as follows. In section 2, we present the state-of-the-art for word embeddings learning. Section 3 presents our approach to word representation using the bag of morphemes. Experiments for intrinsic and extrinsic evaluation are presented and discussed in section 4. We conclude the work in section 5.

\section{Word Embeddings}
This section discusses some methods to learn word embeddings.

Mikolov et al.\cite{word2vec} proposed two architectures to learn word embeddings considering the word context window of every word within the corpus, Skip-Gram and CBOW. The CBOW aims to predict the central word $c$ based on its context, while the Skip-Gram aims to predict the context of a given central word $c$.

Formally, the Skip-Gram goal is to maximize the function $E_{sk}$:

\begin{equation}
E_{sk} = \frac{1}{T}\sum_{t = 1}^{T}\sum_{-c \leq j \geq c, j \neq 0} \mathbf{log} p (w_{t + j} | w_{t})
\end{equation}

where $c$ is the context window size. Due the computation burden of $p (w_{t + j} | w_{t})$, the authors propose the use of Negative Sample \cite{word2vec} and approximate $ \mathbf{log} p (w_{t + j} | w_{t})$ according to equation \ref{skipgram}.
 
\begin{equation}
        \mathbf{log}\sigma(w_{t+j}^{T}w_{t}) + \sum_{i=1}^{k}E_{w_{t} \sim P_{n}(w)}[\mathbf{log}\sigma(-w_{i}^{T}w_{t})]
        \label{skipgram}
\end{equation}

Equation \ref{skipgram}, thus, aims to distinguish the target word $w_{t+j}$ taken from a noise distribution $P_{n}(w)$, where $k$ is the number of negative samples for each target.

Pennington et al. \cite{pennington2014glove} proposed the GloVe model. In contrast to Skip-Gram and CBOW, GloVe uses the corpus global statistics to learn the word embeddings. Let $X$ be a co-occurrence matrix among the words in the corpus, where $X_{ij}$ indicates how many times the word $j$ appear in word $i$ context. The GloVe objective is thus to minimize the following cost function:

\begin{equation}
J = \sum_{i, j = 1}^{V}f(X_{ij})(w_{i}^{T}{\tilde w_{j}} + b_{i} + {\tilde b_{j}} - log(X_{ij}) )^{2}
\label{glove}
\end{equation}

where $V$ is the vocabulary size, $w_{i}$ is the word embedding of the central word $i$, and $\tilde w_{j}$ is the word embedding of the context word $j$. As shown in equation \ref{glove}, the GloVe aim is to approximate the inner product between vectors $w_{i}$ and $w_{j}$ to log the co-occurrence between them. Santos et al. \cite{flavioglove} presents a complementary method to GloVe, whose objective is to use the Paraphrase Dataset \cite{ppdb} to complete the $X$ co-occurrence matrix.

Bojanowski et al. \cite{bojanowski2016enriching} proposed a new approach based on Skip-Gram, where each word works as a set of characters n-grams. Every character n-gram is associated with a continuous vector representation. Thereby, each word is a sum of all character n-gram representations. 

Aavraham et al. \cite{avraham2017interplay} present a study on how different types of linguistic information (surface form, lemma, morphological tag) affect the semantic and syntactic relation of word embeddings. The authors consider three sets of information: $W$, $L$, and $M$. $W$ is a set of the word surface forms, $L$ is the set of lemma and $M$ a set of morphological tags. Each word $i$ in the vocabulary $V$ is a vector $v_i$ added the W, L, and M representations. After obtaining this new representation, the authors use Skip-Gram to learn the word embeddings.

Cotterell et al. \cite{cotterell2019morphological} proposed an extension of the Log-Bilinear model to learn word embeddings. The proposed variation, called Morpho-LBL, consists of adding a multi-objective to the LBL model to optimize the model so that it predicts the next word and its respective morpheme.

ELMo \cite{peters2018deep} is a reference model for characters n-gram. Their representations are computed by two independent LSTM \cite{hochreiter1997long}: one analyzes the context on the right (step forward) and the other analyzes the context on the left (step backward). The final word representation consists of the weighted sum between all the intermediate representations.

BERT \cite{devlin2018bert} is currently the most prominent word embedding model. It uses attention mechanism \cite{vaswani2017attention} to learn the word contextual representation, a characteristic that allows the identification of which of the contextual information is the most relevant for final word vector representation. Unlike traditional word embeddings and ELMo technique, BERT considers the order of neighboring words on the left and right simultaneously. Similarly to ELMo, words can have different vector representations depending on the context in which they are inserted.

A bunch of works considers the addition of morphological knowledge into the model. Luong et al. \cite{luong2013better}, for instance, incorporate explicitly morphological knowledge in word embeddings. Each word $i$ has a vector $v_i$ and its morpheme vectors. After obtained the new representations of each word, the method uses the Natural Language Model \cite{nlm} to learn the word embeddings. Such kind of incorporation is also the case of Qiu et al. \cite{qiu2014co}. In the first step, they perform morphological analysis of each word $i$ in vocabulary $V$. Thus, each word $i$ is a sum of its vector $v_w$ and its morpheme vectors. The authors use CBOW to learn the word and morpheme representations. It is essential to highlight that the authors gave a weight of $0.8$ to vector $v_i$ and $0.2$ to the morpheme vectors. Xu et al. \cite{xu2017implicitly} also consider morphological knowledge addition in the word embeddings but unlike the works cited so far, this is done implicitly. Each word $i$ is a vector $v_i$ and the set $M_i$, which corresponds to the morpheme meanings. To build $M_i$, first, the authors extract the suffix, prefix, and root of the word $i$ and adds to the set $M_i$. After obtaining the whole new words representations of a vocabulary $V$, they use the CBOW method to learn the word embeddings. Qiu et al. \cite{qiu2014co} proposed an approach to incorporate morphological knowledge into the CBOW architecture where each word is represented by a token uniquely identifying it and the tokens of its morphemes. Similarly, \cite{salama2018morphological} incorporates the words morphological category during word embeddings training; they use CBOW as the base model. Dalvi et al. \cite{dalvi2017understanding} add morphological knowledge into the neural machine translation connection. The authors investigate which part of the decoder is best to add morphological knowledge.

Our proposal is very much related to these latter approaches.

\section{Morphological Skip-Gram}\label{subsec:method}
The proposed approach aims to incorporate morphological knowledge in the Skip-Gram method. We first present the morphological analysis and the baseline model FastText. Next, we detail our proposed method Morphological Skip-Gram (MSG). Lastly, we point out the differences between FastText and MSG.

\subsection{Morphological Analysis} \label{sec:morf_analysis}
Morphological analysis is the task of finding the morphemes of a word. The morpheme is the smallest unit that carries the meaning of a word. The morphemes obtained by the morphological analysis are classified in (i) \textbf{Radical/base}: part common to a certain set of words, from which other words will be formed; (ii) \textbf{Gender and number ending}: has the function of indicating whether the word is in masculine or feminine, plural or singular; (iii) \textbf{Thematic vowel}: links the radical to the endings (terminal elements indicative of the inflections of words) that form the words, constituting the theme; (iv) \textbf{Affixes (prefixes and suffixes)}: prefixes are the particles that are located before the radical and suffixes appear afterward; 

\subsection{FastText}
The Skip-Gram method uses a different vector to each word and ignores information about the inner word structure. The FastText model proposes a score function considering the inner word structure information. In FastText, each word $w$ is represented as a bag of n-grams characters. The full word $w$ is also added in the bag allowing the model to learn the continuous vectorial representations of the words and their characters n-grams. Using the Portuguese word $federativas$ as an example, the 4-gram and 5-gram characters generate the following bag of tokens:

\begin{center}
    <fede, eder, dera, erat, rati, ativ, tiva, ivas, feder, edera, derat, erati, rativ, ativa, tivas>, <federativas>
\end{center}

In practice, the FastText model uses the characters n-grams of size 3, 4, 5, and 6.

Supposing that we have a dictionary $G$ of characters n-grams, where $|G| = K$. Given the word $w$, we denote $G_w \subset G$ as the set of characters n-gram present in $w$, and each n-gram g in $G$ is associated with a continuous vectorial representation named as $z_g$. Thus, the word $w$ is represented by the sum of all vectorial representation of its characters n-gram ($z_g$). Finally, we obtain the following score function used to predict when the word $c$ appear in the context of $w$:

\begin{equation}
    s(w, c) = \sum_{z_g \in G_w} z_g^{T}v_c
\end{equation}

Rewriting the objective function of Skip-Gram using the FastText score function, we obtain:

\begin{equation}
E_{fasttext} = \frac{1}{T}\sum_{t = 1}^{T}\sum_{-c \leq j \leq c, j \neq 0} \mathbf{log} (p (w_{t + j} | w_{t}))
\end{equation}
 
Where $\mathbf{log} p (w_{t + j} | w_{t})$ is calculated through:

\begin{equation}
\begin{multlined}
    \mathbf{log} (p (w_{t + j} | w_{t})) = log(\sigma(s(w_{t+j}, w_{t}))) \\ + \sum_{i=1}^{k}E_{k_{t} \sim P_{n}(w)}[log(\sigma(-s(w_{i}, w_{t})))]
\end{multlined}
\end{equation}

As shown above, word representation considers the character's n-gram representation. As a consequence, the new FastText score function also incorporates the context word information in the vectorial representation of the characters n-gram and consequently in the considered words.

\subsection{Morphological Skip-Gram}
Our proposal uses the morphology of the words as part of the full word representation. The aim of the Morphological Segmentation Task is to segment words in its morphemes, the smallest meaning-carrying unit of a word. For each word $v$ in the vocabulary $V$, we define a set $m_{v}$, where:

\begin{equation}
    m_{v} = \{x \mid x \textrm{ is  a  morpheme of v} \}
\end{equation}

We used Morfessor toolkit \cite{smit2014morfessor} to build the set $m_{v}$. The Morfessor is a popular toolkit for statistical morphological segmentation. It is composed of a family of unsupervised learning methods. 

In the original Skip-Gram, each word $v$ is represented by only one vector $w_v$. In our proposal, each word is represented as:

\begin{equation}
    k_v = w_v + \sum_{x}^{m_{v}}z_{x}
    \label{kv}
\end{equation}

Where $z_{x}$ is the vectorial representation of each morpheme $x$ of word $v$. Thus, as in Skip-Gram, the MSG also has two continuous representations for each word. The first one considers the word as the center of the sentence. The second representation of the word is the context. 

The MSG aim is to maximize the function $E_{msg}$:

\begin{equation}
E_{msg} = \frac{1}{T}\sum_{t = 1}^{T}\sum_{-c \leq j \geq c, j \neq 0} \mathbf{log} p (k_{t + j} | k_{t})
\end{equation}

We also use Negative Samples through training optimizing $\mathbf{log} p (k_{t + j} | k_{t})$ according to equation \ref{msg}.

\begin{equation}
    log\sigma(k_{t+j}^{T}k_{t}) + \sum_{i=1}^{k}E_{k_{t} \sim P_{n}(w)}[log\sigma(-k_{i}^{T}k_{t})]
    \label{msg}
\end{equation}

As we discussed before, some methods such as GloVe \cite{glove}, Skip-Gram \cite{word2vec}, CBOW \cite{word2vec} present interesting results, but they do not use the inner structure word information. Our proposed method, MSG, overcomes this limitation adding morphological word knowledge into the Skip-Gram architecture. Besides, every word morpheme token ($z_x$ in equation \ref{kv}) has equal weight to the word token ($m_v$ in equation \ref{kv}), differently from approach present in \cite{xu2017implicitly} that add morphological knowledge into CBOW architecture and give different weights to morpheme vectors and word vectors.

\subsection{Differences between the FastText and the Morphological Skip-Gram}
\begin{figure*}
\centering
\includegraphics[scale=0.5]{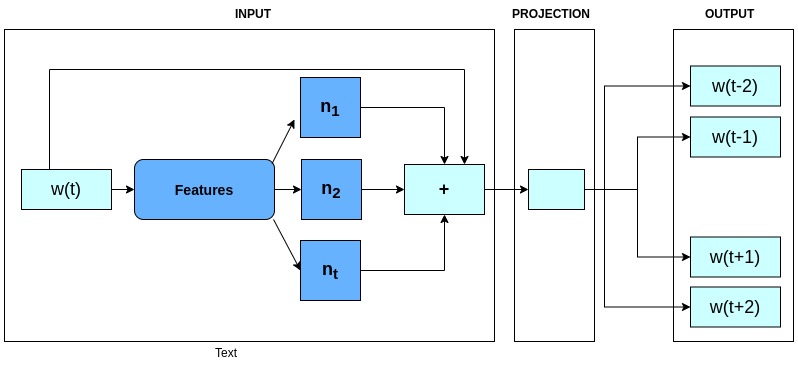}
\caption{FastText and MSG general model}
\label{fig:fasttext-msg}
\end{figure*}

The FastText method is our baseline; thus, this section will focus on discussing the difference between our proposed method and it. Figure \ref{fig:fasttext-msg} present a visual representation of the Morphological Skip-Gram and FastText architectures in a general way. In order to simplify the figure, we use only a context of size 2 (the output, thus, has only four words): $w(t-2)$, $w(t-1)$, $w(t+1)$ and $w(t+2)$, two previous and two posterior words related to the central word $w(t)$. 

The main difference between the two models is the source of information added in the Skip-Gram architecture, represented in figure \ref{fig:fasttext-msg} as a feature block. The feature block output in MSG is all word morphemes, while in FastText are all characters n-grams. An important point to highlight is the complexity of the two models, which is dependent on the number of tokens produced in the feature block. In the training step of both, it is necessary to compute the derivatives concerning each one of vectorial representations (morphemes in the Morphological Skip-Gram and characters n-grams in the FastText). Thus, in the example of the Portuguese word $federativas$, using the sets of 4-gram and 5-gram of characters, we need to learn 16 representations in the FastText model. However, using the set $\{federa, tiva, s\}$ obtained from the morphological analysis, we need to learn four representations in the MSG model. That example shows up the huge difference between uses morphological knowledge as inner structure information and all character n-grams as inner structure information.

Yet from the perspective of the architecture, we can make an analogy with an artificial neural network composed of three layers: input, projection, and output. The projection layer represents the word embeddings, morphemes, or n-grams; the more substantial dimension in this layer results in more learned parameters and hence a larger space to represent information.

\section{Evaluation and discussion}

We performed intrinsic evaluation to compare our proposed model with the baseline FastText. All the intrinsic evaluation methods used are universal assessment methods well known by the scientific community.

Intrinsic evaluation measure the word embeddings quality compared to human judgment. We used 15 well know datasets to perform the intrinsic evaluation. The datasets fall into three categories: (i) Similarity, (ii) Analogy, and (iii) Categorization.

\begin{itemize}
     \item \textbf{Similarity}: datasets composed of pairs of words, where each pair has an average rank defined by humans. This category consists of 6 datasets: SimLex999 \cite{hill2015simlex}, MEN \cite{bruni2014multimodal}, WordSimilarity353 \cite{finkelstein2001placing}, Rare Words \cite{finkelstein2001placing}, RG65 \cite{rg65} e Turk \cite{mturk}.
     \item \textbf{Analogy}: datasets composed of two pairs of words relative to a specific relation. For example, (man, woman), (king, queen). This category is composed of 3 datasets: Google \cite{mikolov2013efficient}, MSR \cite{msr}, SemEval 2012.2 \cite{c2}.
     \item \textbf{Categorization}: problems involving a sentence or word with a binary target. This category is composed by 3 datasets: AP \cite{ap}, BATTING \cite{batting}, BLESS \cite{bless}.
\end{itemize}

We use the toolkit developed by \cite{jastrzebski2017evaluate} to evaluate the word embeddings in these 15 datasets.

\subsubsection{Corpora and training details}
The corpus 1 billion words language model benchmark \cite{chelba2013one} was used to train all the word embeddings. This corpus consists of approximately 1 billion words. We do not perform the pre-processing step because the corpus is ready to use.

\begin{table}
    \centering
    \caption{Training details}
    \label{tab:trainingDetails}
    \begin{tabular}{ll}
        \toprule
        \textbf{PARAMETERS} & \textbf{VALUES} \\
        \midrule
        dimensions & 50, 100, 200, 300 \\ 
        iterations & 20 \\ 
        threads & 12  \\
        negative sample  & 5 \\
        context window  & 5 \\
        learning rate & 0.1 \\
        \bottomrule
    \end{tabular}
\end{table}

Table \ref{tab:trainingDetails} shows the parameter values used along with the training of all word embeddings. It is important to note that the values of threads, negative samples, context window, and learning rate were chosen according to the default values of the FastText project.

\subsubsection{Results and discussion}
In these experiments, we only compare MSG with FastText because, in FastText paper, the authors already made a comparison between FastText and other word embeddings models and achieved better or competitive results. There are recent deep learning-based models that learn good word representations, such as BERT and ElMo, but BERT and ELMo are context-sensitive models; the same word has different representations depending on its context. However, FastText, GloVe, Word2Vec, Morphological Skim-Gram are context insensitive because, after training, the word has the same representation no matter its context. Thus,  these word embeddings models are in different categories, and we do not consider a fair comparison. 

In experiments, we used a Dell desktop with the 8th generation Intel core i5 processor, 8GB of RAM, 1TB of storage, running the Ubuntu 16.04 operating system. We ensure that both models are executed in the same environment situation to obtain the values of Table \ref{tab:time}. 

Table \ref{tab:metricas} presents a summary of all used datasets in the intrinsic evaluation, highlighting the dataset, category, and used metric. High values represent better results.

\begin{table}
    \centering
    \caption{Summary of datasets and metrics}
    \begin{tabular}{lll}
        \toprule
        Dataset & Category & Metrics \\
        \midrule
        SimLe999                     & Similarity      &  $\rho-Spearman$ \\
        MEN                          & Similarity      &  $\rho-Spearman$ \\
        Word Similarity 353          & Similarity      &  $\rho-Spearman$ \\
        Rare Words                   & Similarity      &  $\rho-Spearman$ \\
        RG 5                         & Similarity      &  $\rho-Spearman$ \\
        Turk                         & Similarity      &  $\rho-Spearman$ \\
        Google                       & Analogy         &  Accuracy \\
        MSR                          & Analogy         &  Accuracy \\
        SemEval 2012.2               & Analogy         &  Accuracy \\
        AP                           & Categorization  &  Purity \\
        BLESS                        & Categorization  &  Purity \\
        BATTING                      & Categorization  &  Purity \\
        \bottomrule
    \end{tabular}
    \label{tab:metricas}
\end{table}

Tables \ref{tab:analogia}, \ref{tab:similaridade}, and \ref{tab:categorizacao} present the results of the experiments performed using tasks of analogy, similarity and categorization, respectively. Table \ref{tab:time} shows a comparison of training time between the models. The name ft-d50 means that these word embeddings were trained using FastText (FT) and have dimension 50. While msg-d50 means it was trained using the Morphological Skip-Gram (MSG) and has dimension 50.

\begin{table}
    \caption{Analogy results}
    \centering
    \begin{tabular}{llll}
    \toprule
    \textbf{Name} & \textbf{Google} & \textbf{MSR} & \textbf{SE-2012} \\
    \midrule
    ft-d50 & 0.000 &    0.001 &    0.12747 \\ 
    msg-d50 & 0.005 &     0.001    & 0.128 \\
    ft-d100 & 0.008 &    0.004    & 0.143 \\
    msg-d100 & 0.064 &    0.0126    & 0.144  \\
    ft-d200 & 0.087&    0.051    & 0.155   \\ 
    msg-d200 & 0.242 &    0.104    & 0.164  \\
    ft-d300 & 0.128 &    0.082    & 0.168 \\
    msg-d300 & 0.332    & 0.211 &    0.180  \\
    \bottomrule
    \end{tabular}
    \label{tab:analogia}
\end{table}

Considering the Analogy results, the MSG model presented superior results to the FastText in all datasets and all embeddings dimensions. The except case was the dataset MSR using embeddings ft-d50 and msg-d50, in which both had the same performance (Table \ref{tab:analogia}). However, embeddings of dimension 50 did not have a good performance on all three datasets. We observed a variation of the results with the change of the size of the embedding. For instance, in the results using the Google dataset, the msg-d50 model obtained 0.005 while the msg-d300 obtained 0.332, in other words, the size of the embeddings has much influence on the model performance at Google Analogy benchmark.

\begin{table}
    \caption{Similarity Results}
    \centering
    \begin{tabular}{lllllll}
        \toprule
        \textbf{Name} & \textbf{SL9} & \textbf{MEN} & \textbf{WS353} & \textbf{RW} & \textbf{RG65} & \textbf{Turk}\\
        \midrule
        ft-d50    & 0.25 & 0.64 & 0.61 & 0.25 & 0.58 & 0.63 \\ 
        msg-d50  & 0.25 & 0.64 & 0.61 & 0.23 & 0.56 & 0.63 \\
        ft-d100   & 0.29 & 0.69 & 0.64 & 0.28 & 0.65 & 0.66 \\
        msg-d100 & 0.30 & 0.68 & 0.64 & 0.25 & 0.64 & 0.63 \\
        ft-d200   & 0.33 & 0.71 & 0.65 & 0.30 & 0.71 & 0.67 \\
        msg-d200 & 0.33 & 0.71 & 0.65 & 0.29 & 0.67 & 0.64 \\
        ft-d300   & 0.34 & 0.73 & 0.66 & 0.31 & 0.73 & 0.67 \\
        msg-d300 & 0.35 & 0.71 & 0.66 & 0.29 & 0.68 & 0.66 \\
        \bottomrule     
    \end{tabular}
    \label{tab:similaridade}
\end{table}

From table \ref{tab:similaridade}, we can see the MSG model presented competitive results compared to FT. The methods showed a difference of a maximum $0.05$ (RG65 dataset using the ft-d300 and msg-d300 models). It is essential to point out that the FT was the best in all the Rare Words (RW) dataset cases, generalizing better in the unknown words scenario. It is important to highlight that FastText is designed to deal with word out of vocabulary, rare words, and word with spelling errors. This FastText characteristic is due to characters n-grams because even if a word is out of vocabulary, some (or all) of its characters n-grams can be in n-grams vocabulary. Thus, since the RW scenario is composed of words with low frequency, some of its characters n-grams can be present in word with high frequency.

\begin{table}
    \caption{Categorization Results}
    \centering
    \begin{tabular}{llll}
        \toprule
        \textbf{Name} & \textbf{AP} & \textbf{BLESS} & \textbf{BATTING}\\
        \midrule
        ft-d50 & 0.61 &    0.75 &    0.43 \\ 
        msg-d50 & 0.62    & 0.74 &    0.42 \\ 
        ft-d100 & 0.60    & 0.77    & 0.43 \\
        msg-d100 & 0.59 &    0.77 &    0.45\\
        ft-d200 & 0.61 &    0.80 &    0.43 \\
        msg-d200 & 0.60 & 0.81 &    0.43 \\
        ft-d300 & 0.59    & 0.81    & 0.43  \\
        msg-d300 & 0.60    & 0.82 &0.42  \\
        \bottomrule
    \end{tabular}
    \label{tab:categorizacao}
\end{table}

Considering the Table \ref{tab:categorizacao} results, both models presented similar performance, differentiating by a maximum of $0.01$. 

\begin{table}
    \caption{Training Time}
    \centering
    \begin{tabular}{ll}
        \toprule
        \textbf{Name} & \textbf{AP} \\
        \midrule
        ft-d50 & 307m27  \\ 
        msg-d50 & 196m11 \\ 
        ft-d100 & 409m42  \\
        msg-d100 & 268m50 \\
        ft-d200 & 647m32  \\
        msg-d200 & 414m30 \\
        ft-d300 & 853m58 \\
        msg-d300 & 541m35    \\
        \bottomrule
    \end{tabular}
    \label{tab:time}
\end{table}

Considering Tables \ref{tab:analogia}-\ref{tab:categorizacao} results, the MSG model is better than the FT in the analogy tasks, whereas, in the Similarity and Categorization evaluations, the models had comparable performance. A possible explanation of why the MSG model presents better results than FT in analogy task is that some tokens of the FT brute force (all characters n-gram) solution may be adding noise in the word representation. Thus, since the morphological analysis only adds expert linguistic knowledge, it produces only the necessary tokens (morphemes) to represent the inner word structure.

Table \ref{tab:time} shows that the training time of the MSG model is approximately $40\%$ faster than the FT model. That improvement in training time is expected because the MSG model uses expert knowledge information to represent the inner structure of the word, unlike FastText, which uses a force brute solution. 

From the results obtained in intrinsic evaluation, we can see that Morphological Skip-Gram achieved competitive results compared with FastText and is approximately $40\%$ faster than it. Both methods have the Skip-Gram as base architecture, being different on the inner structure information: MSG uses morphemes of the word, and FastText uses all characters n-grams. Thus, there is strong evidence that the morphological information is sufficient to represent the inner structure word information, and the FastText brute force solution has some characters n-gram, which are not useful.

\subsection{Morphological Skip-Gram Limitations}
The intrinsic evaluation shows that our proposed method, MSG, presents competitive results and is about $40\%$ faster than our baseline FastText. However, even presenting excellent results in intrinsic evaluations, the MSG has limitations. For instance, we only consider the Skip-Gram architecture to add morphological knowledge; there are other architectures, such as GloVe and Neural Language Model, that we can introduce morphological knowledge. Besides, there is other syntactic expert knowledge to be added in word embeddings, such as stemming, radical, and lemma.

\section{Conclusion}

This work presented the Morphological Skip-Gram (MSG) method to learn word embeddings using expert knowledge, especially the morphological knowledge of words. The MSG uses the morphological structure of words to replace the n-grams bag of characters used in the FastText model. The use of such a bag of n-grams is a brute force solution as it tries all possible combinations of characters n-grams of a word. As the purpose of using this bag of n-grams is also to learn the information about the internal structure of the words, we consider the morphological information more robust and informative. We compared MSG with FastText in 12 benchmarks. Results show that MSG is competitive compared to FastText and takes 40\% less processor time than FastText to train the word embeddings. Keeping the quality of word embeddings and decreasing training time is very important because usually, a corpus to training embeddings is composed of 1B tokens. For example, the Common Crawl corpora contain 820B tokens.

As future works, we intend to conduct experiments using different tools for morphological analysis to evaluate their impact on the MSG. Furthermore, we intend to investigate other sources of semantic and syntactic knowledge to incorporate in word embeddings; some of those can be part-of-speech tagging, stemming, and lemmatization.

\section*{Acknowledgments}

The authors thank CAPES and FAPITEC-SE for the financial support [Edital CAPES/FAPITEC/SE No 11/2016 - PROEF, Processo 88887.160994/2017-00] and LCAD-UFS for providing a cluster for the execution of the experiments. The authors also thank FAPITEC-SE for granting a graduate scholarship to Flávio Santos, and CNPq for giving a productivity scholarship to Hendrik Macedo [DT-II, Processo 310446/2014-7].

\bibliographystyle{unsrt}

\end{document}